\documentclass[
]{ceurart}

\usepackage{listings}
\usepackage{subcaption}
\usepackage{booktabs}
\usepackage{bm}
\usepackage[justification=centering]{caption}
\begin{document}

\copyrightyear{2022}
\copyrightclause{Copyright for this paper by its authors.
  Use permitted under Creative Commons License Attribution 4.0
  International (CC BY 4.0).}

\conference{Woodstock'22: Symposium on the irreproducible science,
  June 07--11, 2022, Woodstock, NY}

\title{Programmable Cellular Automata}

\author[1,3]{Ahmed Khalifa}[%
orcid=0000-0002-7839-9432,
email=ahmed@akhalifa.com,
url=http://www.akhalifa.com,
]
\cormark[1]
\address[1]{Institute of Digital Games, University of Malta, 20 L. Esperanto, L-Imsida, Malta}

\author[2]{Muhammad Umair Nasir}[%
orcid=0000-0002-2458-9599,
email=muhammad.nasir@wits.ac.za,
]
\cormark[1]
\address[2]{University of the Witwatersrand, 1 Jan Smuts Ave, Braamfontein, Johannesburg, 2017, South Africa}

\author[3]{Matthew Siper}[%
orcid=0009-0008-6570-8209,
email=m@thenof1.com,
]
\address[3]{Nof1, 665 Broadway, Brooklyn, NY 10012, USA}

\author[2]{Steve James}[%
orcid=0000-0003-4366-4125,
email=steven.james@wits.ac.az,
url=https://sdjames.me,
]

\author[3,4]{Julian Togelius}[%
orcid=0000-0003-3128-4598,
email=julian@togelius.com,
url=http://julian.togelius.com,
]
\address[4]{Game Innovation Lab, New York University, 370 Jay St, Brooklyn, NY 11201, USA}

\cortext[1]{Corresponding author.}
% \fntext[1]{These authors contributed equally.}

\begin{abstract}
  Cellular automata is a local computation paradigm where complex behavior can arise from local interactions between simple functions. This paradigm has been used to explain many systems such as biological processes, traffic simulation, computer networks, etc. In games, cellular automata have been used in games such as SimCity and for the generation of spatial content such as caves or dungeons. However, creating effective local rules is hard and unintuitive. Cellular automata can be effectively evolved, but may still be hard to interpret. In this work, we introduce the concept of programmable cellular automata, where we represent the system as Python code. We also modularize the cellular automata into local functions and a decision function. Local functions take a local neighborhood and return a value, while the decision function takes the output of the local functions and decides the value of the next state. Separating the cellular automata into modules written in Python helps with understanding how these systems are working. We also explore adding global functions where they take the whole state and compute a function from it. We tested generating levels for three different games from the PCG Benchmark. The results showed that global functions decrease the number of iterations that cellular automata need to solve a problem, and that we cannot find solutions for some problems with purely local functions. Looking into the generated functions, we can see common functions that have been used in different experiments, which not only helps us understand the generator but also helps us understand these games better and what is important for them.
\end{abstract}


\maketitle

\section{Introduction}

The appeal of cellular automata is that simple local computation can create complex global phenomena. A single cellular automaton implements a very simple function, mapping the surroundings of a cell to the next state of the cell. But applying this function iteratively to every cell in a two- (or three-, or n-) dimensional lattice can yield surprisingly complex behavior. Most famously, the Game of Life~\cite{conway1970conway} features an impressive menagerie of gliders, factories, blinkers, spaceships, and so on. The expressive powers of these $2^18$ possible automata are indeed surprising. Perhaps even more surprising, it has been shown that the Game of Life is Turing complete, meaning that any algorithm can be implemented within a sufficiently large lattice of this automaton. The power of these simple automata has inspired some rather fringe theories, in particular Stephen Wolfram's idea that the universe itself is a cellular automaton~\cite{wolfram2002science}; this idea has inspired fiction such as Greg Egan's Permutation City~\cite{egan1994permutation}.

More prosaically, cellular automata have proven useful tools for many tasks. CA have been used to model biological and chemical growth processes~\cite{ermentrout1993cellular,kier2000cellular}, and to simulate, e.g., traffic~\cite{nagel1992cellular} or computer networks~\cite{ren2002cellular}. Within games, cellular automata are one of the core abstractions within the SimCity game series~\cite{wright2003lessons}. They are also useful for level generation, where it has been shown that they can generate, for example, organic-looking cave systems~\cite{johnson2010cellular}.

The canonical CA is a purely discrete affair and can be visualized simply as a table, making the structure very easy to understand and edit for humans. However, many more expressive function spaces are possible. In particular, a relatively recent variant is the Neural Cellular Automaton (NCA)~\cite{mordvintsev2020growing}. In an NCA, the state of a cell at the next time step is decided by a weighted sum of the neighboring cells passed through a nonlinear function. In other words, the cell acts as a neuron. An NCA can be seen as a single-layer convolutional neural network looped on itself.

While NCAs are significantly more expressive, at least per cell, than regular cellular automata, they are equally significantly harder to understand. Much like how a neural network is harder to understand than a lookup table.

Are there other ways of representing cellular automata that retain (or even increase) the expressivity from NCAs but are easier for humans to inspect and author? Yes. Program code. We can simply represent the functions in a modern high-level programming language such as Python. If the only input to the function is the state of the cells around the target cell and the output is the state of the cell at $t+1$, this is still cellular automata, yet one that can be understood by anyone who can read the programming language.

In many cases, you want to learn or automatically construct your cellular automaton rather than building it by hand, in particular if you want to rapidly obtain a CA that fulfils a particular purpose, such as a level generator for a new game. If the new use case comes with an evaluation function (or verifiable reward, as it is also called), one can use an evolutionary algorithm to search for suitable automata. This works with all CA representations. However, with a programmatic representation, we can make use of the tremendous power of code-writing LLMs~\cite{chen2021evaluating}. Agentic coding with prompts to improve the CA can work as a replacement for or complement to traditional mutation and recombination operators.

Another benefit of a programmatic representation is the potential for increased modularity. One could define a cellular automaton as not one but several functions that connect in a specified way, for example, with the output of one as input to another. That way, it is easy to understand, improve, and swap the functions separately. This increased modularity also allows us to include global functions. This is of interest because whereas there exist Turing-complete cellular automata (such as the Game of Life, see above), it seems to be very hard in practice to accomplish certain tasks with entirely local computation, particularly within limited time frames. For example, in many cases you may need a certain number (such as 1, 10, or more than 5) of cells with a particular activation; this requires some type of counter mechanism, which is fiendishly hard and inefficient to implement using only local computation.

In this paper, we describe an initial exploration of programmable cellular automata (PCA). We propose an architecture for PCAs which combines multiple local functions with, optionally, global functions that provide information about the current state and location that get fed to a decision function that decides the values for the next state.

\section{Related work}

Research in cellular automata and procedural content generation (PCG) focused on extending the algorithm capabilities to generate more complex content such as Minecraft buildings~\cite{green2019organic}, dungeon levels~\cite{green2019two,hidayat2024application}, spatial layouts~\cite{ng2026procedural,ng2025hexagonal}, terrain~\cite{macedo2017improving,wu2021procedural}, city layouts~\cite{temuccin2020using}, gameplay mechanic~\cite{cook2016towards}, etc. This direction is mostly common in game development since the generated algorithms are constructive, so they are well suited for online generation~\cite{short2017procedural}. 

Research in cellular automata has focused on evolving the rules such that they can compute some global value~\cite{mitchell1996evolving,mitchell1994evolving}. This is due to the difficulty of designing local rulesets that can compute these global values. In PCG, this is about combining two paradigms: constructive approaches (cellular automata)~\cite{shaker2016constructive} and search-based approaches~\cite{togelius2011search}. In some of the work~\cite{mahlmann2012spicing,adams2017interactive}, they evolved the cellular automata rules and starting state, and the number of iterations to generate levels or maps for video games. Evolving the rules, the starting state, and the number of iterations together means that the evolution is evolving an indirect representation of a single level instead of a generator. Others looked at cellular automata as a generator~\cite{pech2016game,ashlock2013landscape,adams2017procedural} and evolved the rules of the level generator for top-downgames. Earle et al.~\cite{earle2022illuminating} expanded the rule evolution by representing the rules as a neural cellular automata (NCA) and evolved it using a quality diversity algorithm and tested it on multiple games. Earle and Togelius~\cite{earle2025autoverse} evolved NCAs that represent game mechanics for new environments to train RL agents. Sudhakaran et al.~\cite{sudhakaran2021growing} evolved NCA to generate 3D voxel assets.

With the rise of LLMs~\cite{brown2020language}, procedural content generation research focused on using LLMs as a generator to generate content~\cite{gallotta2024large} such as level generators~\cite{sudhakaran2023mariogpt,todd2023level,huang2025word2minecraft,jiang2026agentic}, game generators~\cite{todd2024gavel,hu2024game,nasir2026mortar}, story generators~\cite{nasir2024word2world}, etc. Another direction is that, instead of generating content or the solution, you generate code that can solve the problem. This is done with the help of a search/evolutionary algorithm. Romera et al.~\cite{romera2024mathematical} used the LLM to evolve a scoring function to rank solutions such that the system can select the best candidate solution. Lehman et al.~\cite{lehman2023evolution} evolved code that generates 2D robot designs in an open-ended evolution fashion~\cite{taylor2016open}. Novikov et al.~\cite{novikov2025alphaevolve} used a quality-diversity algorithm with an LLM to evolve code that can solve multiple different problems. The LLM managed to write more efficient matrix multiplication code than the ones that have been used so far. Ma et al.~\cite{ma2024eureka} evolved a reward function for RL algorithms and tested it on 29 different environments, where in a lot of cases the trained RL agent with these evolved functions beats human-trained agents.

This paradigm of writing code is still new, but it works perfectly with the PCG domain, as the LLM agents can write a fast constructive generator that can be used directly in games. Siper et al.~\cite{siper2026procedural} used an LLM to write Python code as a constructive level generator~\cite{shaker2016constructive} for 4 different games. They combined the LLM's ability to write code with an evolutionary loop to find a generator that can generate consistent levels that pass the playability criteria. It also used the LLM to abstract and summarize the generated code into utility functions that other chromosomes can use to improve quickly. This work is inspired by their work, as we look at CA as modular pieces where each piece can be represented by a function that can be produced using any type of code generation, including LLMs.

\section{Programmable Cellular Automata}
\begin{figure*}
    \centering
    \includegraphics[width=\linewidth]{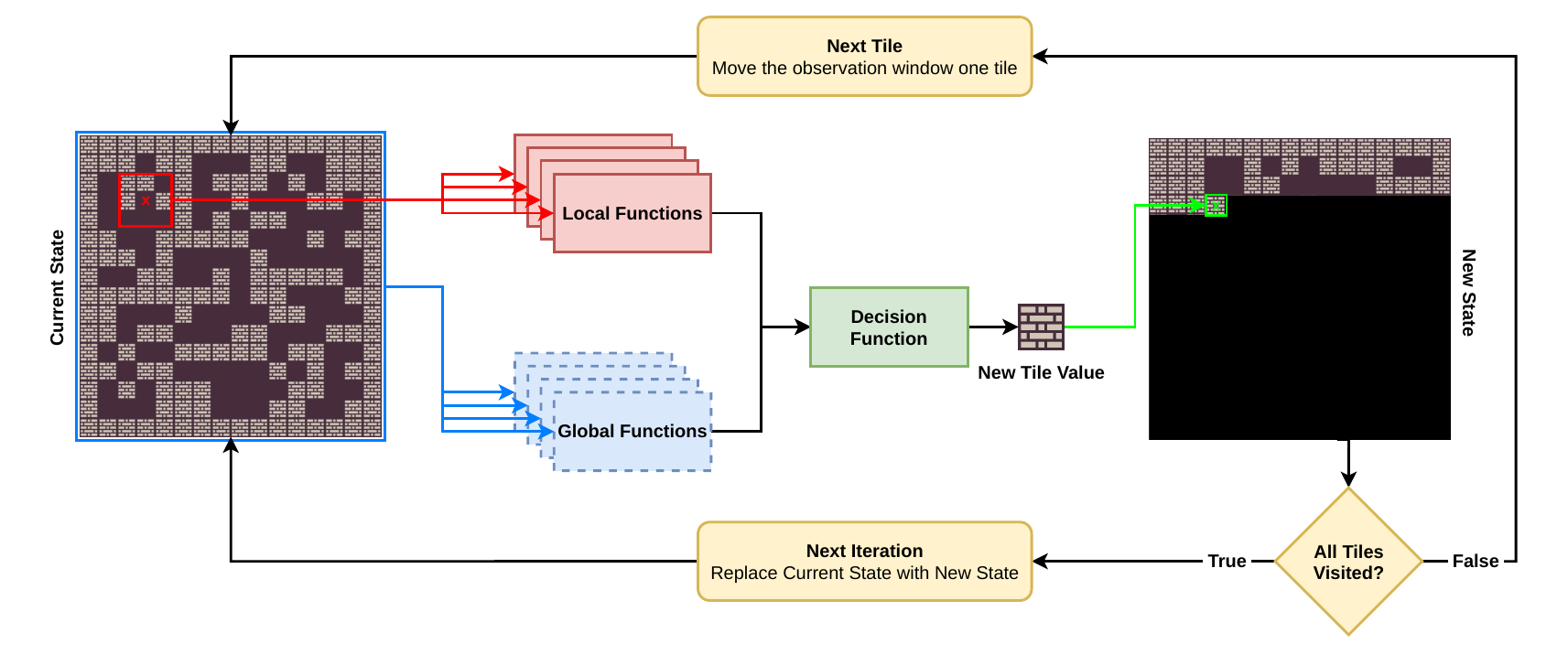}
    \caption{Programmable Cellular Automata follows the same loop as cellular automata, where the current state gets processed one tile at a time and produces a new state; then, in the end, the new state becomes the current state, and you repeat again. The difference can be seen in the middle part where we divided into two or three types programs: Local Functions (compute a function based on local observation), Global Functions (Optional: this is not traditional CA as it takes a full observation and compute a function), and finally Decision function (compute the final value for the current location in the new state using all the output from the other functions).}
    \label{fig:pca}
\end{figure*}

Programmable Cellular Automata (PCA) is a new formulation of Cellular Automata (CA). Instead of defining the CA as a table of local transitions, it is represented as a program. In this respect, it is similar to Neural Cellular Automata (NCA)~\cite{mordvintsev2020growing}. The difference is that, in NCA, all the local functions are represented as convolutional kernels, while the decision function is represented as a fully connected layer. This allows CA to be more expressive but, at the same time, not understandable in terms of its rules. PCA is a more general formulation of this where all these functions are represented as programs (such as program trees, Python code, neural networks, etc). 

Figure~\ref{fig:pca} shows PCA, where at each step the system sends the local observation to all the local functions (instead of the convolutional kernels), and then the decision function (instead of the linear layer) takes all these local computations and decides on the cell value. The PCA iterates over all the cells and computes the new value to be part of the next state. After the PCA finishes iterating over all the cells, the next state replaces the current state of the whole lattice and a new iteration starts.

Although the CA paradigm is focuses on local computation, we are interested in extending it to include global functions and see their effect. The global functions take the whole state instead of a small local observation and try to compute a helping value. In this case, the decision function, at each step, takes the result of all the local computations (from the local functions) and, if available, the global state computations (from the global functions) and decides on the next cell value. This version of PCA is not fully local computationally, but has capabilities to do more complex operations within a small number of iterations.

There are a few constraints for PCA on all these programs:
\begin{itemize}
    \item \textbf{Local Functions:} should take some form of local neighborhood (such as Von Neumann or Moore neighborhood) from the current state and produce a value(s). 
    \item \textbf{Global Functions (Optional):} should take the whole current state and produce a value(s).
    \item \textbf{Decision Function:} should take all the different values from the local functions and global functions (if existing) and return a valid value(s) for the next cell state.
\end{itemize}
The most important thing is that the local/global functions shouldn't change anything in the state, and the decision function shouldn't take anything from the state itself; instead, it should take the computed values from these functions.

Finally, CA are usually \emph{Synchronous}, which means that the changes in each cell only depend on the current state and only replace the current state with new values when everything is being computed. We retain this convention, but we note that global functions could benefit from being \emph{Asynchronous}. \emph{Asynchronous} means that any modification by the decision function to a cell will affect the next computation of the global function. For example, imagine you want a function that counts the number of players, and we are using a \emph{Synchronous} global function; if there are no players, the value will be 0 for all the local interactions, which makes the decision function add a player for each local location. If it is \emph{Asynchronous}, after execution at each local location, the system can update the global count so the next location will know that there is a player has already been added.

\section{Automated Discovery of Programmable Cellular Automata}
Instead of writing programmable cellular automata as humans, we can automate the process of writing them utilizing genetic programming~\cite{poli2008field,koza1990genetic}. In this section, we go over the different parts that are needed to allow genetic programming/evolutionary algorithms to explore new PCAs.

\subsection{Representation and Initialization}
The chromosome is an array of $l$ local functions, $g$ global functions, and an decision function, where $l$ is the number of local functions (at least $1$) and $g$ is the number of global functions, which can be $0$ if you decide not to have any global functions. The functions can be represented as any type of program, it can be represented as trees, strings that represent Python code, or machine learning programs (like neural networks). In this work, we focus on programs represented as Python code because it is easy to understand, and because it makes it easy to utilize current Large Language Models (LLMs) to generate better initial functions than random generation of trees or random neural networks, which can speed the search.

\subsection{Variational Operators}\label{sec:operators}
For variational operators, programmable cellular automata allow for crossover and mutation operators. Because the chromosome is represented as an array of programs, we can apply crossover easily using uniform crossover, one-point crossover, or other types. In this project, we are using uniform crossover where each program can be selected from one of the parents with equal probability. Of course, swapping a local function or global function might not have the same effect as swapping the decision function (since the decision function affects the state changes). Exploring the effect of that is out of scope of this work.

For mutation, we only need a way to generate a new function; this can be random sampling of the code tree, or using an LLM to generate new code, or adding some noise to neural network weights, etc. Since, in the current work, we represent the functions as Python code, we are using an LLM to generate new functions. This can happen in two ways: 
\begin{itemize}
    \item \textbf{Random Sampling:} We prompt the LLM to generate the selected function (local, global, or decision function) given its restrictions (input parameters and/or output) and use the output directly from the LLM. In principle, we would not need to use an LLM at all for this, but any type of code generation as long as it can generate code that respects the given constraints.
    \item \textbf{Context Sampling:} Similar to the above, but besides the restrictions of the function, we also give it context about the rest of the functions (local, global, and decision functions) that it works with and the fitness score that it achieves using these functions. This is harder to achieve without an LLM. Using context helps to guide the LLM towards generating different functions that could improve the fitness of the overall chromosome.
\end{itemize}

\subsection{Fitness}
For evolving the PCA, we need to evaluate its output on multiple inputs. We ran the PCA $n$ times on multiple different inputs. For each run, we ran the PCA for $m$ iterations, and if during these iterations it found a solution, we stopped; if not, we evaluated all the states using our fitness function and returned the maximum value. So the fitness function for each PCA is defined by equation~\ref{eq:fitness}.
\begin{equation}\label{eq:fitness}
    f = \begin{cases}
        1 + \frac{1}{n}\Sigma_{i=0}^{n} D_{i} & \text{if } \forall_i \in \{0, 1, .., n\}, Q_i = 1\\
        \frac{1}{n}\Sigma_{i=0}^{n} Q_{i} & otherwise
    \end{cases}
\end{equation}
\begin{equation}\label{eq:quality}
    Q_i = \max_{k \in m}(q_k)
\end{equation}
\begin{equation}\label{eq:diversity}
    D_i = \min_{k \in n}(dist(i, k))
\end{equation}
where $f$ is the final fitness, $D_i$ is the minimum distance of all other outputs from the different runs (as defined in equation~\ref{eq:diversity}), $Q_i$ is the highest reached quality of each output from the runs over its iterations (as defined in equation~\ref{eq:quality}), $q_k$ is the quality of the state $k$ in run $i$, and $dist(i, k)$ is a distance function between two outputs from the PCA.

The fitness function integrates not just the quality ($Q_i$) but also the diversity of the output content ($D_i$). It uses diversity as a cascaded fitness~\cite{togelius2007towards}, which means that if the output is never fully functional in all $n$ runs, diversity won't contribute to the fitness. Diversity in a lot of problems is very important since we want our PCA to respond differently to different inputs, but if that is not the case (where the more important thing is the solution being just functional), then you can remove that term from equation~\ref{eq:fitness}.

\section{Experiments}

\begin{figure}
    \centering
    \begin{subfigure}[b]{0.33\textwidth}
        \centering
        \includegraphics[height=100pt]{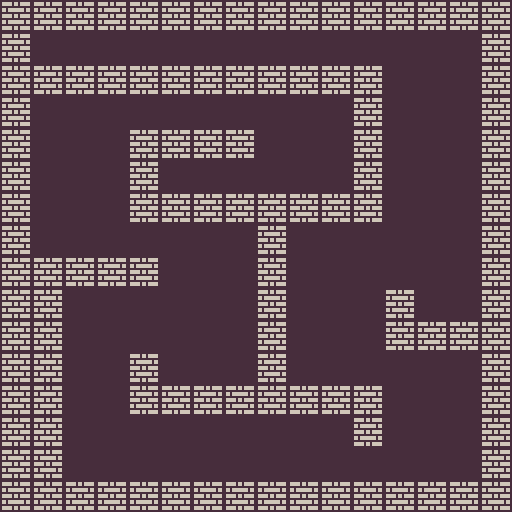}
        \caption{Binary}
        \label{fig:domains_binary}
    \end{subfigure}
    \hfill % Adds horizontal space between subfigures
    \begin{subfigure}[b]{0.33\textwidth}
        \centering
        \includegraphics[height=100pt]{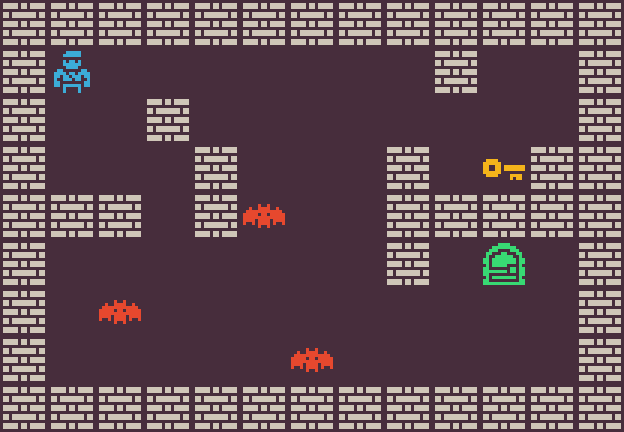}
        \caption{Zelda}
        \label{fig:domains_zelda}
    \end{subfigure}
    \hfill % Adds horizontal space between subfigures
    \begin{subfigure}[b]{0.33\textwidth}
        \centering
        \includegraphics[height=100pt]{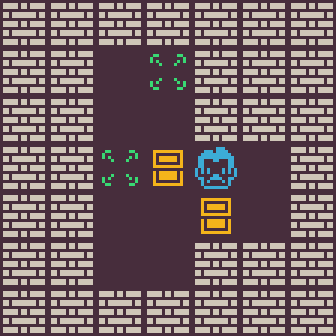}
        \caption{Sokoban}
        \label{fig:domains_sokoban}
    \end{subfigure}
    % \hfill % Adds horizontal space between subfigures
    % \begin{subfigure}[b]{0.26\textwidth}
    %     \centering
    %     \includegraphics[height=80pt]{images/example_loderunner.png}
    %     \caption{Lode Runner}
    %     \label{fig:domains_loderunner}
    % \end{subfigure}
    \caption{Example of human-designed levels for all the 3 game domains. Each of these levels satisfies the corresponding quality metric.}
    \label{fig:domains}
\end{figure}

In this paper, we test PCA on the domain of Games. Each PCA represents a level generator~\cite{johnson2010cellular} for different problems from the PCG Benchmark~\cite{khalifa2025procedural}. For each of the problems, we use the quality and distance metrics provided by the benchmark. Figure~\ref{fig:domains} shows the selected problems from the PCG Benchmark, which are:
\begin{itemize}
    \item \textbf{Binary} is a simple 14x14 (without borders) maze made of solid and empty tiles where the final maze needs to be fully connected and has a path of at least 28 tiles (as shown in figure~\ref{fig:domains_binary}).
    \item \textbf{Zelda} is a simple 11x7 (without borders) dungeon-crawling game inspired by The Legend of Zelda (Nintendo, 1986) dungeon rooms. The goal of the player is to grab a key and reach a door without getting killed by enemies. The levels need to have at least 1 player, 1 key, 1 door, and 3 enemies. The solution length (getting the key then getting the door) is at least 18 moves (as shown in figure~\ref{fig:domains_zelda}).
    \item \textbf{Sokoban} is a simple 5x5 (without borders) puzzle game based on a Japanese puzzle game with the same name (Thinking Rabbit, 1981). The goal of the game is for the player to push all the crates onto the targets. The levels need to have 1 player, at least 1 crate, and an equal number of crates and targets. The solution length (pushing all the crates to the target using an A* solver) is at least 10 moves to avoid generating extremely simple levels but also not complex enough that the solver can't solve it (as shown in figure~\ref{fig:domains_sokoban}).
    % \item \textbf{Lode Runner} is a 32x21 puzzle platformer game based on a game by the same name (Brøderbund, 1983). The goal of the game is to collect all the gold without being caught by the guards in the level. The trick is that the player can't jump, it can move left and right, climb ladders, slide on ropes, and dig holes in the floor to either trap enemies or fall through them. The levels need to have 1 player, 6 gold, and 3 enemies. The level should be solvable (the player can collect all the gold) and at least be 20\% explorable and have a similar structure to the original game (as shown in figure~\ref{fig:domains_loderunner}).
\end{itemize}

For all the experiments, we are using a Moore Neighborhood with a size of $3x3$ for local observation. For our local functions, we are using \emph{Synchronous} cellular automata for our PCA, while the global function is \emph{Asynchronous}. We ran all the evolved PCAs for 100 iterations, stopping when they reach a goal state. For mutation, we are using Claude 4.8 Opus with a maximum of 4096 tokens (to force small functions). Since we are using an LLM for mutation, we use Context Sampling mentioned in section~\ref{sec:operators}. For evaluation, we evaluate each PCA on $30$ random fixed starting inputs to avoid randomness affecting evaluation while still having enough samples to evaluate that PCA. Finally, for the evolution, we are using a population of $40$ with elitism of 10\% that we run for $50$ generations. Selection uses tournament selection of size $7$, and we are using uniform mutation with a rate of $10\%$ (which means each function has a $10\%$ chance of being replaced) and uniform crossover with equal probability between both parents.

We test two hyperparameters to understand their effect on the behavior of the PCA. We test the effect of different numbers of local functions ($l = \{1, 5, 10, 50\}$) and the different numbers of global functions ($g = \{0, 1, 5\}$). A total of $36\ (3 \text{ (number of problems)} \times 4 \text{ (number of local functions)} \times 3 \text{ (number of global functions)})$ different experiments; for each one, we run it $5$ times for stability (total runs equal to $180$).

\section{Results}

\begin{figure*}
    \centering
    \includegraphics[width=\linewidth]{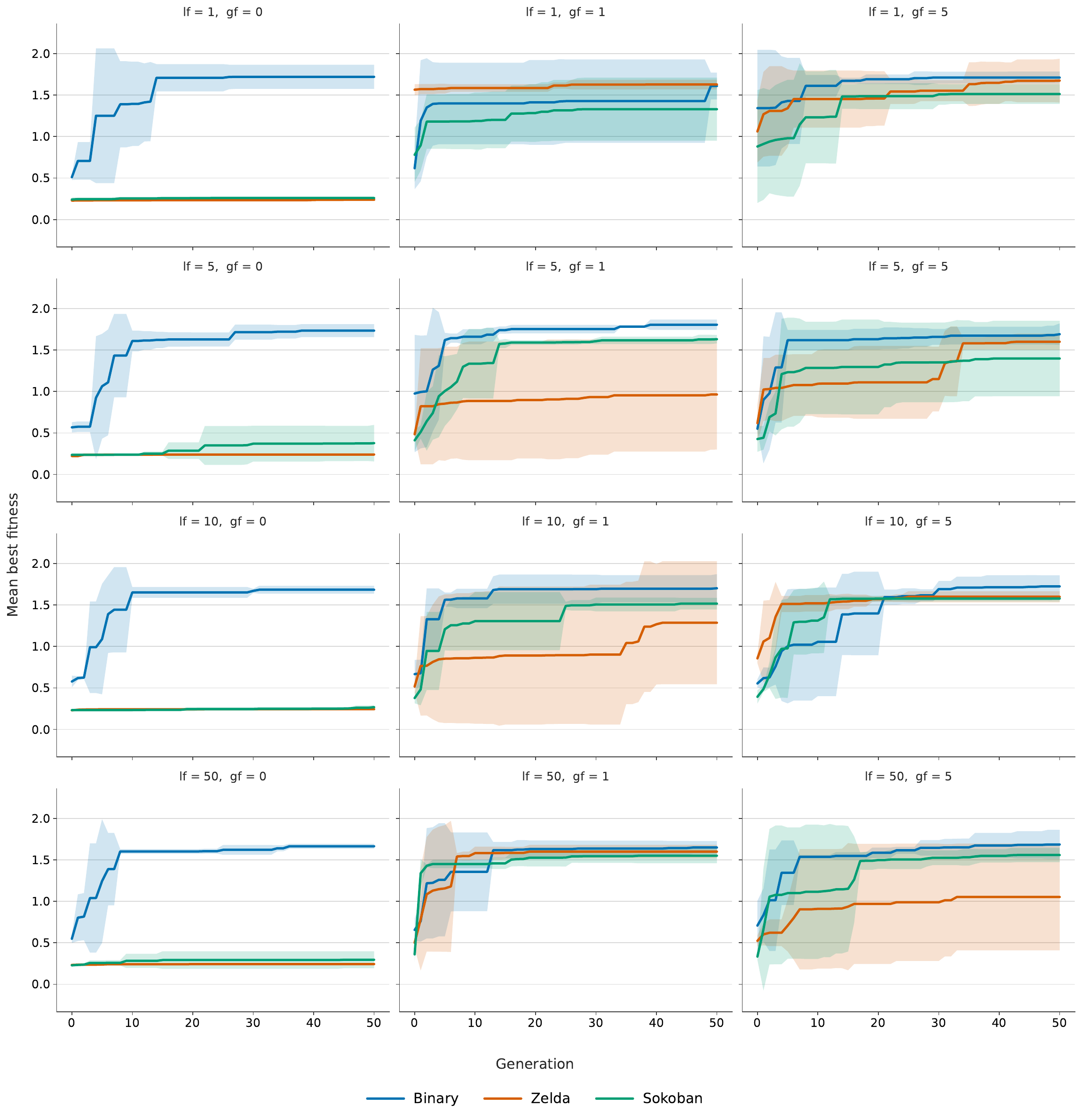}
    \caption{Fitness progression for all three games across the different configurations of the number of local functions and global functions. The error bars are the 95\% confidence interval calculated over the 5 different runs.}
    \label{fig:fitness}
\end{figure*}

Figure~\ref{fig:fitness} shows the fitness improvement across all three domains; we noticed that having at least one global function helps the evolution find a generator that can solve all the problems pretty efficiently. It was surprising that more global functions didn't help as much, but we believe that, due to the simplicity of the problem, as most problems are solved using 1 global function (see table~\ref{tab:playability}), adding more global functions doesn't improve the evolutionary process much.

\begin{table}
    \centering
    \begin{tabular}{|l|c|cccc|}
        \cline{3-6}
        \multicolumn{2}{c|}{} & \multicolumn{4}{c|}{\# Local} \\
        \hline
        Game & \# Global & 1 & 5 & 10 & 50 \\
        \hline
        \hline
        \multirow{3}{*}{Binary} & 0 & $100\% \pm 0\%$ & $100\% \pm 0\%$ & $97\% \pm 3\%$ & $99\% \pm 3\%$ \\
         & 1 & $98\% \pm 2\%$ & $100\% \pm 0\%$ & $100\% \pm 0\%$ & $98 \pm 2$ \\
         & 5 & $98\% \pm 2\%$ & $98\% \pm 1\%$ & $100\% \pm 0\%$ & $99\% \pm 1\%$ \\
        \hline
        \hline
        \multirow{3}{*}{Zelda} & 0 & $0\% \pm 0\%$ & $0\% \pm 0\%$ & $0\% \pm 0\%$ & $0\% \pm 0\%$ \\
         & 1 & $99\% \pm 1\%$ & $46\% \pm 59\%$ & $70\% \pm 65\%$ & $98\% \pm 5\%$ \\
         & 5 & $93\% \pm 10\%$ & $100\% \pm 0\%$ & $99\% \pm 1\%$ & $67\% \pm 34\%$ \\
        \hline
        \hline
        \multirow{3}{*}{Sokoban} & 0 & $2\% \pm 2\%$ & $16\% \pm 32\%$ & $0\% \pm 0\%$ & $11\% \pm 23\%$ \\
         & 1 & $96\% \pm 5\%$ & $96\% \pm 3\%$ & $97\% \pm 5\%$ & $100\% \pm 0\%$ \\
         & 5 & $99\% \pm 1\%$ & $99\% \pm 3\%$ & $97\% \pm 4\%$ & $98\% \pm 3\%$ \\
        \hline
    \end{tabular}
    \caption{The playability percentage of our best generator over 100 sampled levels from each experiment with a 95\% confidence interval for all three game domains.}
    \label{tab:playability}
\end{table}

\begin{table}
    \centering
    \begin{tabular}{|l|c|cccc|}
        \cline{3-6}
        \multicolumn{2}{c|}{} & \multicolumn{4}{c|}{\# Local} \\
        \hline
        Game & \# Global & 1 & 5 & 10 & 50 \\
        \hline
        \hline
        \multirow{3}{*}{Binary} & 0 & $24\% \pm 8\%$ & $26\% \pm 9\%$ & $20\% \pm 9\%$ & $17\% \pm 5\%$ \\
         & 1 & $16\% \pm 13\%$ & $33\% \pm 15\%$ & $23\% \pm 19\%$ & $14\% \pm 7\%$ \\
         & 5 & $27\% \pm 9\%$ & $29\% \pm 12\%$ & $28\% \pm 12\%$ & $23\% \pm 19\%$ \\
        \hline
        \hline
        \multirow{3}{*}{Zelda} & 0 & $27\% \pm 4\%$ & $29\% \pm 0\%$ & $28\% \pm 5\%$ & $36\% \pm 5\%$ \\
         & 1 & $16\% \pm 5\%$ & $25\% \pm 14\%$ & $17\% \pm 5\%$ & $14\% \pm 2\%$ \\
         & 5 & $22\% \pm 29\%$ & $11\% \pm 3\%$ & $14\% \pm 6\%$ & $22\% \pm 12\%$ \\
        \hline
        \hline
        \multirow{3}{*}{Sokoban} & 0 & $2\% \pm 1\%$ & $3\% \pm 2\%$ & $1\% \pm 0\%$ & $4\% \pm 5\%$ \\
         & 1 & $7\% \pm 2\%$ & $12\% \pm 5\%$ & $7\% \pm 4\%$ & $10\% \pm 4\%$ \\
         & 5 & $8\% \pm 3\%$ & $11\% \pm 4\%$ & $10\% \pm 3\%$ & $10\% \pm 4\%$ \\
        \hline
    \end{tabular}
    \caption{The diversity percentage (how many levels are different from each other) of our best generator over 100 sampled levels from each experiment with a 95\% confidence interval for all three game domains.}
    \label{tab:diversity}
\end{table}

\begin{table}
    \centering
    \begin{tabular}{|l|c|cccc|}
        \cline{3-6}
        \multicolumn{2}{c|}{} & \multicolumn{4}{c|}{\# Local} \\
        \hline
        Game & \# Global & 1 & 5 & 10 & 50 \\
        \hline
        \hline
        \multirow{3}{*}{Binary} & 0 & $6.903 \pm 0.934$ & $9.157 \pm 5.777$ & $13.637 \pm 1.032$ & $12.350 \pm 10.980$ \\
         & 1 & $15.587 \pm 7.284$ & $14.710 \pm 6.425$ & $15.593 \pm 2.555$ & $17.873 \pm 14.104$ \\
         & 5 & $24.257 \pm 4.803$ & $7.420 \pm 2.659$ & $16.017 \pm 13.849$ & $15.427 \pm 6.259$ \\
        \hline
        \hline
        \multirow{3}{*}{Zelda} & 0 & $-$ & $-$ & $-$ & $-$ \\
         & 1 & $10.013 \pm 1.966$ & $58.247 \pm 51.779$ & $39.453 \pm 55.971$ & $21.920 \pm 20.695$ \\
         & 5 & $28.493 \pm 2.250$ & $15.543 \pm 15.337$ & $16.240 \pm 3.885$ & $41.020 \pm 28.729$ \\
        \hline
        \hline
        \multirow{3}{*}{Sokoban} & 0 & $98.377 \pm 1.867$ & $89.800 \pm 20.091$ & $99.673 \pm 0.703$ & $91.993 \pm 17.217$ \\
         & 1 & $19.057 \pm 10.468$ & $24.223 \pm 13.613$ & $6.863 \pm 4.806$ & $13.533 \pm 2.588$ \\
         & 5 & $19.853 \pm 15.533$ & $22.063 \pm 13.389$ & $24.093 \pm 18.942$ & $11.730 \pm 3.382$ \\
        \hline
    \end{tabular}
    \caption{The number of iterations for our best generator to find a playable level over 100 sampled levels from each experiment with a 95\% confidence interval for all three game domains.}
    \label{tab:iterations}
\end{table}

Another interesting finding, by looking at table~\ref{tab:iterations}, is that the average number of iterations that the PCA needs to find the best solution decreases when using global functions, except for the binary problem, where the optimum was having just 1 local function. This might be because evolution didn't optimize all the local and global functions well compared to having 1 single function. For example, if you have one global function, evolution will optimize it to be efficient at solving the problem, but if evolution has more than one function, it might split functionality or even have inefficient functions that can solve the same problem. Having inefficient functions might be the cause of having more iterations compared to having 1 global function. 

To explain why having a global function helps in decreasing the number of iterations, imagine you are trying to count the number of player tiles in the game. To do it with local interaction, the local function and execute need to propagate whenever it finds a player tile to the rest of the grid to communicate that there is one in that location. This usually takes at least a number of iterations equal to the board size. This can be seen in work by Earle et al.~\cite{earle2022illuminating,earle2023pathfinding}, where the NCA needs to propagate the information through the board for multiple iterations to compute some global qualities such as path length or connectivity using local computations. Having an easy way to compute these values instead of propagating a lot of information cuts the number of iterations overall. This can be easily seen in Binary since it is an easy problem and can be solved in the allocated time slots.

% Earle et al.~\cite{earle2023pathfinding} further showed that NCAs can be hand-designed or trained to perform pathfinding, effectively running breadth-first search by propagating activations through hidden channels one cell per step. This demonstrates both the expressivity of local computation and its cost: global quantities such as path length or connectivity require a number of iterations that scales with the size of the grid.

\begin{figure*}
    \centering
    \includegraphics[width=\linewidth]{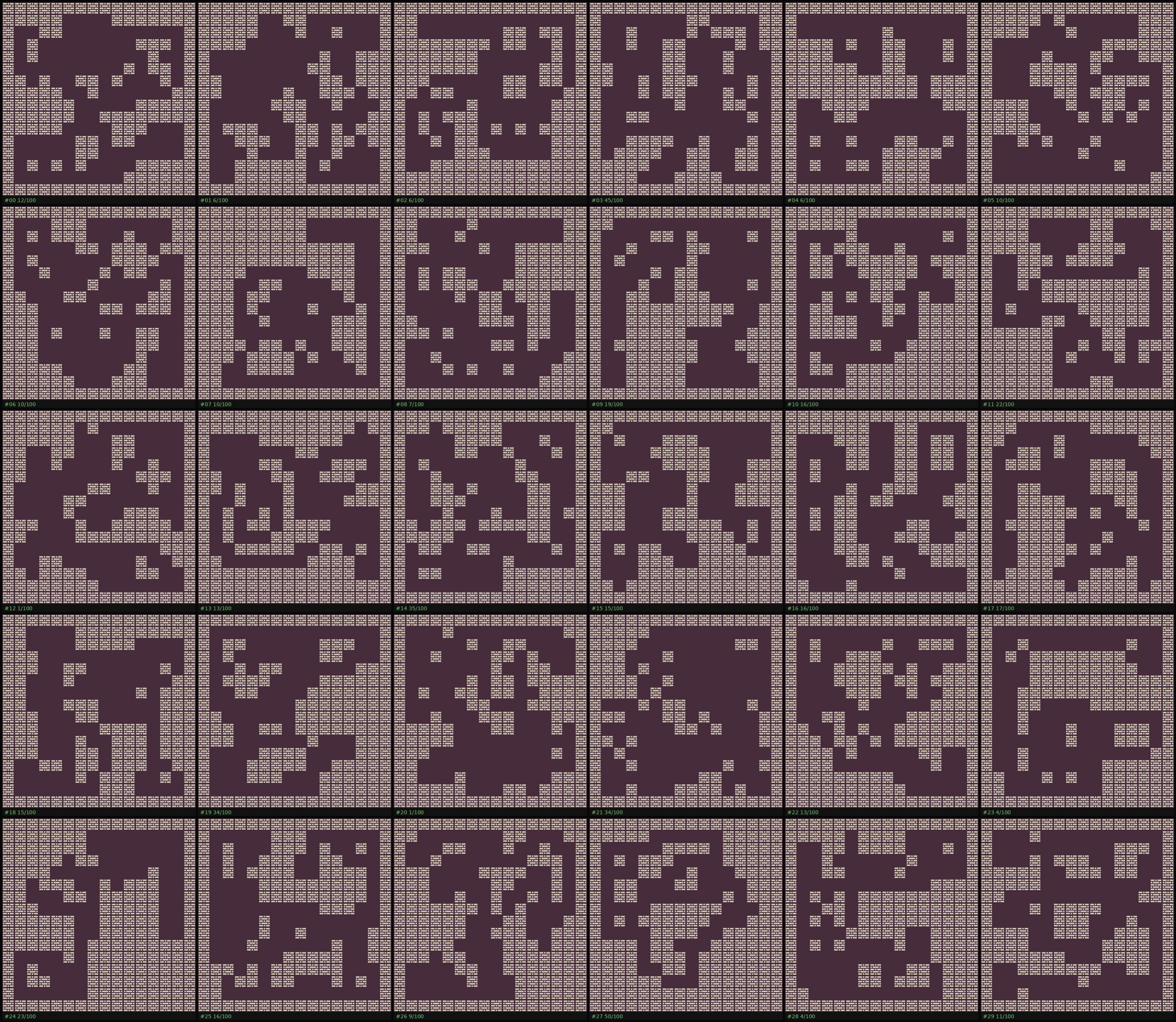}
    \caption{30 generated levels of the Binary domain from our highest-fitness generator, which has 1 local function and 1 global function. The caption states the \# of iterations per generation.}
    \label{fig:levels_binary}
\end{figure*}

\begin{figure*}
    \centering
    \includegraphics[width=\linewidth]{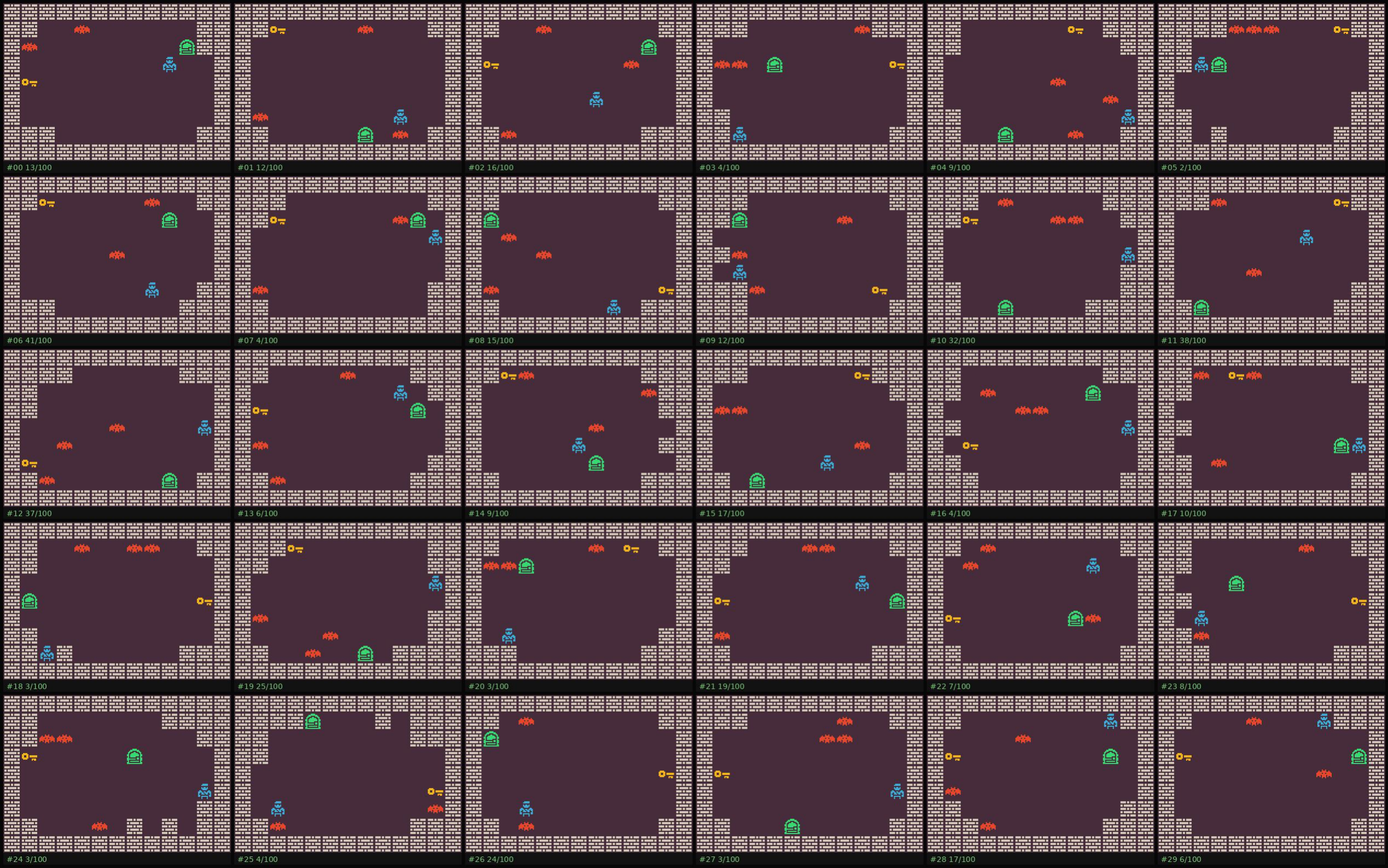}
    \caption{30 generated levels of the Zelda domain from our highest-fitness generator, which has 50 local functions and 1 global function. The caption states the \# of iterations per generation.}
    \label{fig:levels_zelda}
\end{figure*}

\begin{figure*}
    \centering
    \includegraphics[width=\linewidth]{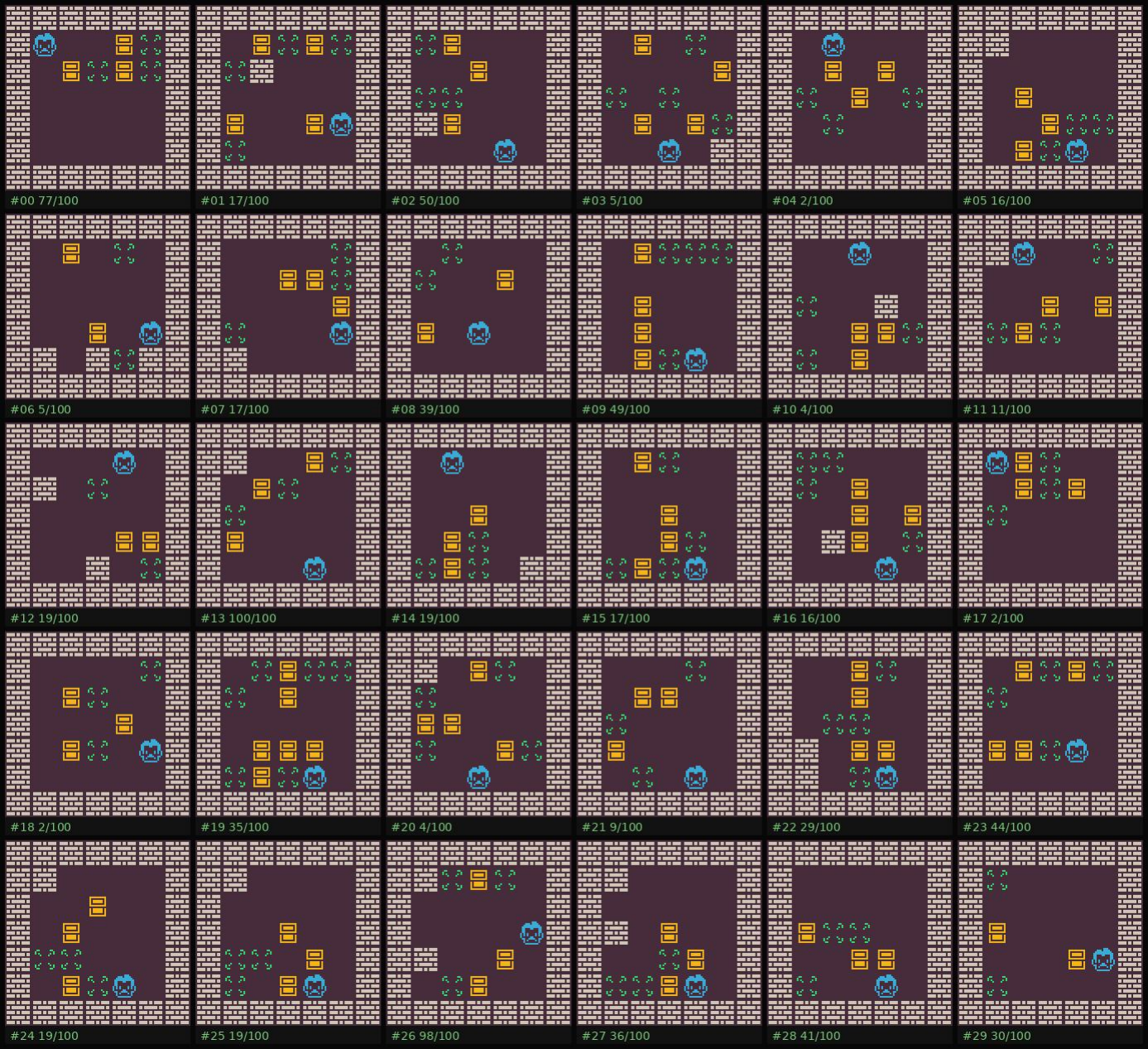}
    \caption{30 generated levels of the Sokoban domain from our highest-fitness generator, which has 10 local functions and 5 global functions. The caption states the \# of iterations per generation.}
    \label{fig:levels_sokoban}
\end{figure*}

We analyze the best generator per domain. Table~\ref{tab:diversity} shows the percentage of levels that the PCG Benchmark labels as passing the diversity criteria. A noticeable result is that although the diversity fitness is high because the fitness (in some cases almost $1.7$), the number of unique levels from each other is low (all under $30\%$). This might be due to how we framed the diversity as part of the fitness function, where we use the distance criteria and not the percentage of different levels. 

We look deeper into the generated content by analyzing the output of the fittest generator from all the runs for all three games, which can be seen in figures~\ref{fig:levels_binary}, ~\ref{fig:levels_zelda}, and ~\ref{fig:levels_sokoban}. The Binary and Sokoban levels look visually diverse and have different solutions. But the Zelda levels don't look as diverse, as the PCA ended up clearing the layout and just moving the player, key, and door in random locations. This is not an issue of the evolved PCA but more of the diversity metric that is being used for Zelda. In Zelda, the diversity is measured based on the shortest path from the player towards the key and from the key to the door, so if you randomize the location of these items, you end up having good diversity.

\begin{table}
    \centering
    \begin{tabular}{|l|p{5cm}|c|ccc|}
        \hline
        Name & Description & \# Lines & Binary & Zelda & Sokoban \\
        \hline
        \hline
        Counting Function & Counts the total number of cells for a given tile. & $5 \pm 3$ & $61\%$ & $61\%$ & $43\%$ \\
        \hline
        Connectivity Function & Calculates the size of the largest connected area for a given tile. & $23 \pm 4$ & $17\%$ & $4\%$ & $19\%$ \\
        \hline
        Border Detector & Counts the number of cells on the borders of the map for a given tile. & $9 \pm 3$ & $0\%$ & $6\%$ & $10\%$ \\
        \hline
        Spread Function & Computes the perimeter of the bounding box enclosing all cells of a given tile. & $11 \pm 4$ & $3\%$ & $6\%$ & $8\%$ \\
        \hline
        Row Distribution Function & Measures how unevenly the tile type is spread across rows. & $10 \pm 4$ & $0\%$ & $2\%$ & $11\%$ \\
        \hline
        Spatial Distribution Function & Rewards tiles on the edge or the center of the map while penalizing corners. & $13 \pm 4$ & $0\%$ & $3\%$ & $7\%$ \\
        \hline
        Clustering Function & Measures how densely clustered the tile type is across the map. & $15 \pm 8$ & $3\%$ & $0\%$ & $5\%$ \\
        \hline
        Distance Function & Computes the maximum Manhattan distance of any pair of tiles in the map. & $12 \pm 4$ & $0\%$ & $2\%$ & $2\%$ \\
        \hline
        Perimeter Function & Counts the total number of edges of tiles that touch either an inactive cell or the map boundary. & $15 \pm 3$ & $6\%$ & $0\%$ & $4\%$ \\
        \hline
        % Weighted Counting Function & Counts the tiles of the given type across the grid, weighting each row by its index so that tiles in lower rows contribute more to the total. & $9 \pm 2$ & $0\%$ & $2\%$ & $2\%$ \\
        Run-Length Function & Returns the maximum horizontal length of connected cells in any row. & $15 \pm 4$ & $8\%$ & $1\%$ & $2\%$ \\
        \hline
        % Dispersion Function & Computes the mean Manhattan distance of active cells from their centroid, giving an integer measure of how spread out the tiles of a given type are. & $12 \pm 2$ & $0\%$ & $1\%$ & $2\%$ \\
        % Row Coverage Function & Counts how many rows contain at least one active tile of the given type, measuring the vertical spread of that tile across the map. & $4 \pm 1$ & $0\%$ & $2\%$ & $0\%$ \\
        % Bounding-Box Function & Computes the horizontal span (width) of the bounding box enclosing all active cells of the given tile type, returning 0 if none are present. & $10 \pm 4$ & $3\%$ & $1\%$ & $1\%$ \\
        % Diagonal Function & It counts the number of set tiles along the main and anti-diagonals, returning twice their difference plus the center tile value as a diagonal-asymmetry feature. & $9 \pm 3$ & $0\%$ & $0\%$ & $2\%$ \\
        % \hline
    \end{tabular}
    \caption{The percentage of time that the top 10 discovered global functions appear in all the experiments and runs (total of 40 runs per game) and the average number of lines for each of these functions.}
    \label{tab:global_functions}
\end{table}

Looking at the global functions discovered during all these runs (table~\ref{tab:global_functions}), we notice that the most common discovered function is a count function. This makes sense since most of these games have constraints on the number of objects, which is an easy global function to discover but very hard to compute using just local functions. We can see that as soon as we have that 1 function in all the domains, it manages to solve the issue (table~\ref{tab:playability}). The second most common function is about connectivity; this is also expected since in most of these games, you want the whole map to be fully connected, as if there are any isolated areas, the level is not considered playable. Finally, the other functions are about the distribution of tiles in the levels; we believe that they use these functions for aesthetics and to make sure the level is well distributed and not clustered in one location.

\begin{table}
    \centering
    \begin{tabular}{|l|p{6cm}|c|ccc|}
        \hline
        Name & Description & \# Lines & Binary & Zelda & Sokoban \\
        \hline
        \hline
        Counting Function & Counts the number of active cells in the local neighborhood. & $5 \pm 2$ & $100\%$ & $85\%$ & $66\%$ \\
        \hline
        Alone Function & Returns whether the center cell is alone with no neighbors. & $10 \pm 4$ & $3\%$ & $24\%$ & $46\%$ \\
        \hline
        Orientation Function & Detect horizontal vs vertical alignment of the active cells in the neighborhood. & $8 \pm 3$ & $36\%$ & $14\%$ & $36\%$ \\
        \hline
        Encoding Function & Encodes the center cell's value with the count of active neighboring cells. & $7 \pm 3$ & $33\%$ & $17\%$ & $27\%$ \\
        \hline
        Vertical Function & Compute the vertical gradient of the neighborhood. & $9 \pm 3$ & $14\%$ & $14\%$ & $26\%$ \\
        \hline
        Weighting Function & Computes a weighted score of the neighborhood: orthogonal neighbors (20), other neighbors (7), and the center cell (1). & $8 \pm 2$ & $3\%$ & $6\%$ & $31\%$ \\
        \hline
        Pattern Function & It compares corner versus edge occupancy, then hashes that difference. & $11 \pm 3$ & $0\%$ & $6\%$ & $24\%$ \\
        \hline
        Corner Function & Computes weighted counts of corner versus edge neighbors. & $10 \pm 3$ & $0\%$ & $5\%$ & $24\%$ \\
        \hline
        % Vertical Gradient Function & Computes the difference between the top-row and bottom-row tile densities in the 3x3 neighborhood, plus the center cell, yielding a signed vertical asymmetry measure. & $7 \pm 2$ & $8\%$ & $11\%$ & $12\%$ \\
        % \hline
        Hashing Function & Computes a hashed scalar for the corner count, center presence, and edge-count parity. & $7 \pm 2$ & $8\%$ & $8\%$ & $11\%$ \\
        \hline
        Contrast Function & Compares the count of set corner cells against set edge cells depending on whether the center cell is set. & $8 \pm 1$ & $0\%$ & $2\%$ & $19\%$ \\
        \hline
        % Corner Detector & Counts how many of the four diagonal corner cells in a 3x3 neighborhood are set, yielding a value from 0 to 4. & $7 \pm 3$ & $11\%$ & $7\%$ & $10\%$ \\
        % Asymmetry Function & It computes a directional asymmetry score by subtracting the bottom-right triangle sum from the top-left triangle sum, biased by twice the center cell's value. & $9 \pm 2$ & $0\%$ & $4\%$ & $15\%$ \\
        % Border Detector & Counts how many active cells of the given tile type lie on the 8-cell perimeter ring of a 3x3 neighborhood, excluding the center cell. & $8 \pm 4$ & $19\%$ & $8\%$ & $5\%$ \\
        % Center Detector & Returns 1 if the center cell of the 3x3 neighborhood is active (non-zero), otherwise 0. & $5 \pm 2$ & $0\%$ & $10\%$ & $4\%$ \\
        % \hline
    \end{tabular}
    \caption{The percentage of time that the top 10 discovered local functions appear in all the experiments and runs (total of 60 runs per game) and the average number of lines for each of these functions.}
    \label{tab:local_functions}
\end{table}

Table~\ref{tab:local_functions} shows the top 10 discovered local functions. The count function is the most common; having this count function transforms the cellular automata into an indirect type, where the decision function decides based on how many tiles surround the center tile. This is pretty common for generating layouts for levels, for example in the work by Johnson et al.~\cite{johnson2010cellular}. We have a big group of pattern-based functions (Orientation Function, Alone Function, Vertical Function, Pattern Function, Corner Function, and Contrast Function); we believe these functions are useful for connectivity and making sure objects are spread well across the map, and the maps are fully connected (that is why they appear with a high percentage in the binary domain). Finally, other functions encode a lot of information that can be used in a full table later to make decisions.

\section{Discussion \& Conclusion}
In this work, we explored the idea of modularizing cellular automata by introducing the idea of programmable cellular automata where each module is considered a function that can be represented by a program. We explored the usage of LLMs to generate these programs and explored the effect of their number on the output results. We expanded the formulation of cellular automata by adding global functions. Although adding global functions is not in the spirit of being fully local, the advantages of adding even 1 asynchronous global function are more than the disadvantages. Adding at least 1 asynchronous global function decreases the number of iterations the cellular automata needs to find the best output, and in some games, it helps the fitness to reach playable levels in a reasonable amount of evolution time.

In our work, the changes are visible in the current state, which means you can always see what the decision function changed, which allows for interesting movement behaviors and patterns~\cite{conway1970conway}. We could expand the idea of state by having a visible part and a hidden part, where the hidden part can act like a memory from the current state to the next state. Doing that won't allow a lot of random changes in the state, and we can make sure the changes are more human-understandable, similar to the hidden states introduced in NCA~\cite{mordvintsev2020growing}. Due to the scope of the paper, this was not tested in the current paper, but this extension might help the local function perform better, as they have more channels to propagate information. 

Another idea is to allow global functions and local functions to return more information. This might help the overall system create more complex functions; for example, the global function can return something like a Dijkstra map, but on the other hand, it can cheat and return the final values, and the decision function just copies it.

Looking at the generated levels, we notice that a lot of the generated levels look visually similar. For example, in Zelda, a generator that generates an empty level and shuffles the position of the objects is enough to pass quality criteria. So instead of fixing the fitness function to incorporate visual diversity, we should use quality-diversity approaches similar to  Earle et al.'s work~\cite{earle2022illuminating} to explore the space of different generators that generate visually distinct levels.

One of the limitations of our implementation is that the local and global functions don't take all the tiles but are only applied to one tile type at a time. This limits the functions; for example, you can't write a BFS algorithm that measures the distance between two different tile types, but you can write functions that measure distance between the same tile type. We restricted the functions to that to make sure the functions are simple, but it also limits the functions that can be produced. It would be interesting to see what will happen when we enable all tile types for all the functions.

An avenue to explore later is to see the effect of the LLM output size; in this work, we allowed the LLM to write code of any size up to 4096 tokens, which is a big size code. Although none of the generated functions were big (as shown in tables~\ref{tab:global_functions} and ~\ref{tab:local_functions}), it would be interesting to explore what will happen if we limit the output token size to something small, such as 128 tokens from the beginning of the evolution. This might force the evolution to be smart and find a better and more efficient solution for these functions overall. We leave this avenue to be explored in future work.

%\begin{acknowledgments}
%  Thanks to Everyone who supported us :D  
%\end{acknowledgments}

%% The declaration on generative AI comes in effect
%% https://ceur-ws.org/GenAI/Policy.html
\section*{Declaration on Generative AI}
% \em Either:}\newline
%   The author(s) have not employed any Generative AI tools.
%   \newline
  
%  \noindent{\em Or (by using the activity taxonomy in ceur-ws.org/genai-tax.html):\newline}
During the preparation of this work, the author(s) used Grammarly in order to check grammar and spelling. We also used Claude Opus to sift through all the generated functions and give them names and descriptions for table~\ref{tab:global_functions} and ~\ref{tab:local_functions}. Finally, Gemini and Claude were used to find missing citations and references in the whole document.
% Further, the author(s) used X-AI-IMG for figures 3 and 4 in order to: Generate images. After using these tool(s)/service(s), the author(s) reviewed and edited the content as needed and take(s) full responsibility for the publication’s content. 

%%
%% Define the bibliography file to be used
\bibliography{sample-ceur}

% \appendix


% The sources for the ceur-art style are available via
% \begin{itemize}
% \item \href{https://github.com/yamadharma/ceurart}{GitHub},
% % \item \href{https://www.overleaf.com/project/5e76702c4acae70001d3bc87}{Overleaf},
% \item
%   \href{https://www.overleaf.com/latex/templates/template-for-submissions-to-ceur-workshop-proceedings-ceur-ws-dot-org/pkfscdkgkhcq}{Overleaf
%     template}.
% \end{itemize}

\end{document}